\documentclass[a4paper]{article}

\usepackage{INTERSPEECH2021}

\usepackage{booktabs}
\usepackage{multirow}
\usepackage{graphicx}
\usepackage{booktabs,tabularx}
\usepackage{siunitx}
\usepackage{ltablex}
\usepackage{tipa} 
\usepackage{stfloats}
\usepackage{lipsum}
\usepackage{enumitem}
\usepackage{hyperref}

\usepackage{array,booktabs,longtable,tabularx,tabulary}

\title{Do Acoustic Word Embeddings Capture Phonological Similarity? \\An Empirical Study}

\name{Badr M. Abdullah$^{1, 2}$, Marius Mosbach$^{1, 2}$, Iuliia Zaitova$^{1, 2}$, Bernd M\"obius$^{2}$, Dietrich Klakow$^{1, 2}$}
\address{
$^{1}$Spoken Language Systems (LSV), Saarland University, Germany \\
$^{2}$Language Science and Technology (LST), Saarland University, Germany \\
Saarland Informatics Campus, Germany
  }
\email{\{babdullah|mmosbach|izaitova|moebius|dietrich\}@lsv.uni-saarland.de}

\begin{document}

\maketitle
\begin{abstract}
Several variants of deep neural networks have been successfully employed for building parametric models that project variable-duration spoken word segments onto fixed-size vector representations, or acoustic word embeddings (AWEs).  However, it remains unclear to what degree we can rely on the distance in the emerging AWE space as an estimate of word-form similarity.  In this paper, we ask: does the distance in the acoustic embedding space correlate with phonological dissimilarity? To answer this question, we empirically investigate the performance of supervised approaches for AWEs with different neural architectures and learning objectives. We train AWE models in controlled settings for two languages (German and Czech) and evaluate the embeddings on two tasks: word discrimination and phonological similarity. Our experiments show that (1) the distance in the embedding space in the best cases only moderately correlates with phonological distance, and (2) improving the performance on the word discrimination task does not necessarily yield models that better reflect word phonological similarity. Our findings highlight the necessity to rethink the current intrinsic evaluations for AWEs. 

\end{abstract}

\section{Introduction}

 Spoken language technologies such as spoken term discovery ~\cite{jansen2010towards, jansen2011efficient,anastasopoulos2017spoken} and query-by-example (QbE) search \cite{zhang2009unsupervised, jansen2012indexing, metze2013spoken}  aim to capture, organize, and facilitate access to the linguistic content of spoken documents while abstracting away from speaker- and context-related sources of variability in speech. To this end, researchers have developed parametric models based on deep neural networks (DNNs) that project variable-length spoken word segments onto speaker-invariant vector representations, known as acoustic word embeddings (AWEs), where acoustic segments of the same word are projected nearby in space \cite{levin2013fixed, bengio2014word, kamper2016deep, settle2016discriminative, settle2017query}. AWEs, and their underlying vector-space acoustic models, enable efficient indexing and retrieval of spoken content at a scale that non-parametric template-based approaches with dynamic programming \cite{heigold2012investigations, de2007template} have failed to deliver.
 
 


Several DNN architectures and learning objectives have been explored in the literature to build AWEs. State-of-the-art AWE models are trained using either contrastive objectives \cite{kamper+etal_icassp16, he+etal_iclr17} or reconstruction objectives \cite{kamper2015unsupervised, kamper2019truly}. AWEs have been used in downstream applications including ASR \cite{bengio+heigold_interspeech14} and QbE search \cite{settle2017query, yuan+etal_interspeech18}. However, evaluating the utility of AWEs using downstream applications is expensive and may not be always feasible. Therefore, researchers have developed an intrinsic evaluation for AWEs based on the acoustic word discrimination task. In this task, AWE models are evaluated based on their ability to determine whether or not two acoustic segments correspond to the same word type \cite{carlin2011rapid, kamper2015unsupervised, kamper2016deep}.

Furthermore, Levin et al. \cite{levin2013fixed} have hypothesized that the distance in the emergent AWE space can be interpreted as a metric of (perceptual) dissimilarity between linguistic units (e.g., phones, syllables, words). However, none of the previous studies has empirically (in)validated this hypothesis with a rigorous evaluation beyond word discrimination. Although previous studies have proposed to incorporate the pronunciation distance in the learning objective  \cite{he+etal_iclr17, yang2019linguistically}, the reported performance showed no improvement on the word discrimination task, while the distance in the AWE space has shown only a weak correlation with orthographic similarity \cite{he+etal_iclr17}. These observations, however, are yet to be systematically investigated across different architectures, objectives, and languages beyond English, which we aim to address in our study.

Since AWE models have been recently adopted as cognitive models of infant phonetic learning \cite{MatusevychSKFG20} and cross-language non-native processing \cite{matusevych2021phonetic}, we argue that more effort should be devoted to analyze and understand the emergent embedding space to make sure it behaves as expected. In this paper, we take a step in this direction and make the following contributions:


\begin{enumerate}[label={(\arabic*)}]
    \item We train AWE models with identical resources and hyperparameters and examine the effects of, and the interplay between, the architecture and learning objective on model performance (\S2).
    
    \item  We analyze the correlation between the distance in the embedding space and word-form (dis)similarity, which we measure using a phonetically-informed extension of Levenshtein distance (\S3 and \S4).
    
    \item We empirically show that while AWE models trained with contrastive objectives outperform other models on the  word discrimination task, they are poor at capturing phonological similarity (\S5). 
\end{enumerate}


\section{Acoustic Word Embedding Models}

\begin{figure*}
    \centering
    \includegraphics[width=0.95\textwidth]{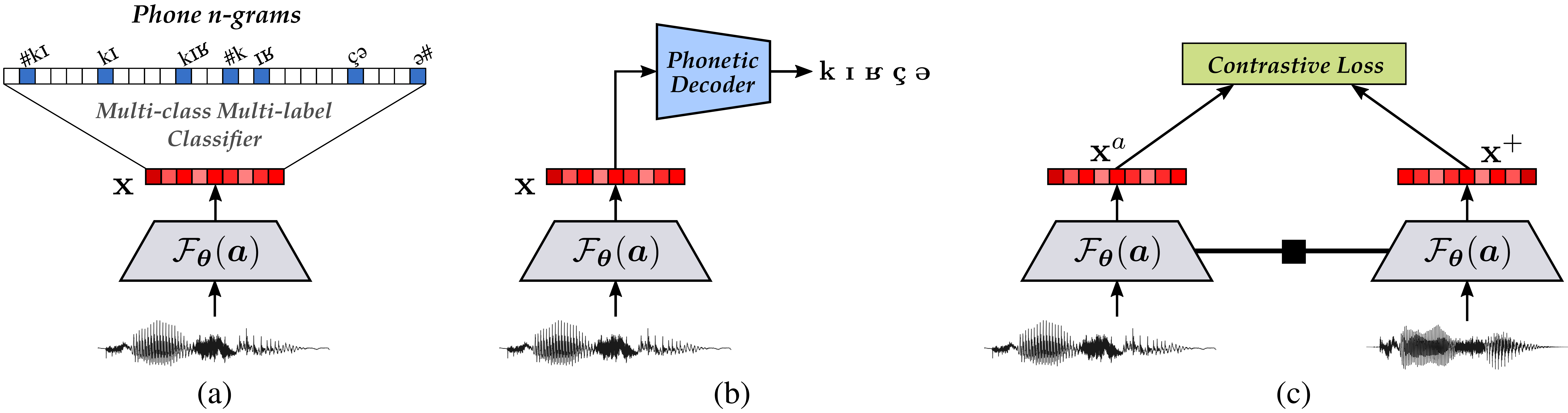}
    \caption{A visual illustration of the learning objectives for the models in our paper: (a) phone $n$-gram detection: a  classification objective based on a multi-class multi-label classifier, (b) word-to-phones: a sequence-to-sequence objective based on decoding the phonological sequence from the acoustic word embedding, and (c) a contrastive siamese objective with a triplet margin loss.}
    \label{fig:models}
\end{figure*}

The core component of a neural AWE model is an acoustic encoder, which can be formally described as a parametric function $\mathcal{F}_{\boldsymbol{\theta}}: \mathcal{A} \xrightarrow[]{} \mathbb{R}^D$, where $\mathcal{A}$ is the (continuous) space of acoustic sequences, $D$ is the dimensionality of the embedding, and  $\boldsymbol{\theta}$ are the parameters of the function. Given an acoustic word segment represented as a temporal sequence of $T$ spectral vectors $\overline{\boldsymbol{a}} = (\boldsymbol{a}_1, \boldsymbol{a}_2, ..., \boldsymbol{a}_T)$, an embedding is computed as $\mathbf{x} = \mathcal{F}_{\boldsymbol{\theta}}(\overline{\boldsymbol{a}}) \in \mathbb{R}^D$. We experiment with convolutional and recurrent architectures for the encoder $\mathcal{F}_{\boldsymbol{\theta}}$ and investigate three learning objectives, which we formally describe below. 

\subsection{Phone $n$-gram Detection Objective}

Following previous work  \cite{settle+etal_icassp19, kamper2020multilingual}, we use a classification objective as our neural baseline (Fig. 1--a). However, we diverge from previous approaches in the classification target of our network. Instead of predicting the word type, we train our model to detect phone sequences that are present in each acoustic segment. For example, consider acoustic segments that correspond to the word type ``\textit{Kirche}''. While previous approaches trained the model to predict the word type as an atomic unit, we first transform the phonetic sequence \textipa{/k I K \c{c} @/} into a set of phone bigrams \{`\#\textipa{k}', `\textipa{kI}', `\textipa{IK}', `\textipa{K\c{c}}',  `\textipa{\c{c}@}', `\textipa{@}\#'\}  and trigrams \{`\#\textipa{kI}', `\textipa{kIK}', `\textipa{IK\c{c}}', `\textipa{K\c{c}@}', `\textipa{\c{c}@}\#'\}. We then train the model to predict the presence of each phone $n$-gram. Formally, given a vocabulary of word types $\mathcal{V}$ where each word is associated with a phonetic sequence $\boldsymbol{\varphi}_{1:N} = (\varphi_1, \dots, \varphi_N)$, we obtain a set of all phone $n$-grams in the training data as $\Phi = \bigcup_{v \in \mathcal{V}}\mathcal{N}(\boldsymbol{\varphi}_{1:N}^v)$, where $\mathcal{N}$ is a function that converts a phone sequence into phone $n$-grams. Each phone $n$-gram in the set $\Phi$ is a prediction target in the network where a multi-class multi-label classification head is connected to the AWE to give the prediction output as ${\mathbf{\hat{y}}} = \sigma(\mathbf{xW} + \mathbf{b}) \in [0, 1]^{|\Phi|}$, where $\sigma$ is a sigmoid activation function, $\mathbf{W}$ is a weight matrix, and $\mathbf{b}$ is a bias vector. The objective is to minimize binary cross-entropy loss  as 
\begin{equation}
\mathcal{L} = -\Big[ \mathbf{y} \cdot \text{log}(\mathbf{\hat{y}}) + (\mathbf{1} - \mathbf{y}) \cdot \text{log}(\mathbf{1} - \mathbf{\hat{y}}) \Big]
\end{equation}
where $\mathbf{y} \in \{0, 1\}^{|\Phi|} $ is the ground truth vector  and $1$ indicates the presence of a phone $n$-gram in the segment and $0$ indicates its absence. This objective has an efficiency advantage over type-based classification approaches since the classification layer is smaller in size, due to the fact that $|\Phi| << |\mathcal{V}|$.


\subsection{Word-To-Phones Objective}

Our second  learning objective is based on sequence-to-sequence learning whereby the network is trained as a word-level acoustic model (Fig. 1--b). Given an acoustic sequence $\overline{\boldsymbol{a}}$ and its corresponding phonetic sequence $\boldsymbol{\varphi}_{1:N}$, the acoustic encoder $\mathcal{F}_{\boldsymbol{\theta}}$ is trained to take as input $\overline{\boldsymbol{a}}$ and produce an AWE $\mathbf{x}$, which is then fed into a recurrent phonetic decoder $\mathcal{G}$ whose goal is to generate the corresponding sequence $\boldsymbol{\varphi}_{1:N}$. The intuition of this objective is that phonologically similar word-forms would usually have overlapping phonetic segments, we thus expect similar words to end up nearby in the embedding space. The objective is to minimize a categorical cross-entropy loss at each timestep in the decoder, which is equivalent to 
\begin{equation}
    \mathcal{L} = -  \sum_{t=1}^{N}{ \text{log } \mathbf{P}_{\theta}(\varphi_{t} | \boldsymbol{\varphi}_{<t}, \mathbf{x})}
\end{equation}
where $\mathbf{P}_{\theta}$ is the probability of the phone $\varphi_{t}$ at timestep $t$, conditioned on the previous phone sequence $\boldsymbol{\varphi}_{<t}$ and the AWE $\mathbf{x}$. 
\subsection{Contrastive Siamese Objective}
The third objective we investigate in this paper is the siamese contrastive objective \cite{bromley1994signature}, which has been extensively explored in the literature with different underlying architectures \cite{settle+livescu_slt16, kamper+etal_icassp16}.  This objective differs from the first two objectives in two aspects: (1) it explicitly minimizes/maximizes relative distances between AWEs of the same/different word types, and (2) models are trained solely on sensory input without symbolic grounding since each word segment is paired with another segment of the same word type. Given a matching pair of AWEs $(\mathbf{x}^a, \mathbf{x}^+)$, the objective is then to minimize a triplet margin loss
\begin{equation}
    \mathcal{L} = \text{max} \Big[0, \mu + d(\mathbf{x}^a, \mathbf{x}^+) - d(\mathbf{x}^a, \mathbf{x}^-) \Big]
\end{equation}
where $\mathbf{x}^-$ is an AWE that corresponds to a different word type, which serves as a negative sample, and $d: \mathbb{R}^D \times \mathbb{R}^D \rightarrow [0, 1]$ is the cosine distance. This objective aims to map acoustic segments of the same word type closer in the embedding space while pushing away segments that correspond to other word types by a distance defined by the margin hyperparameter $\mu$. We experiment with two different strategies for choosing the negative sample  $\mathbf{x}^-$: (1) we randomly sample a negative AWE from the mini-batch, and (2) we choose the segment that minimizes the distance $d(\mathbf{x}^a, \mathbf{x}^-)$, which is known as semi-hard negative sampling \cite{jansen2018unsupervised}. In \S5, we show that the negative sampling strategy has a significant impact on model performance.

\section{Phonological Similarity Measure}

\begin{table}[]
\centering
\caption{Examples of  pairwise word distances with Levenshtein distance (LD) and the phonologically weighted LD (PWLD).}
\label{tab:PWLD}
\begin{tabular}{@{}cclcccc@{}}
\toprule
\multicolumn{2}{c}{\textsc{Word I}} &  & \multicolumn{2}{c}{\textsc{Word II}} & \multirow{2}{*}{LD} & \multirow{2}{*}{PWLD} \\ \cmidrule(r){1-2} \cmidrule(lr){4-5}
Orth. & IPA &  & Orth. & IPA &  &  \\ \midrule
\multirow{4}{*}{\textit{sicher}} & \multirow{4}{*}{\textipa{/z \textsc{I} \c{c} {\textturna}/}} &  & \textit{Becher} & \textipa{/b E \c{c} {\textturna}/} & 2 & 0.263 \\
 &  &  & \textit{Fischer} & \textipa{/f \textsc{I} S {\textturna}/} & 2 & 0.368 \\
 &  &  & \textit{Lichter} & \textipa{/l \textsc{I} \c{c} t {\textturna}/} & 2 & 0.632 \\
 &  &  & \textit{sitzt} & \textipa{/z \textsc{I} ts t/} & 2 & 0.795 \\ \bottomrule
\end{tabular}
\end{table}

We adopt the phonologically weighted Levenshtein distance (PWLD)  as our measure of phonological distance, or similarity, between different word-forms \cite{fontan2016using}. The PWLD metric extends the string-based Levenshtein distance (LD) by conditioning the cost of phone substitutions on phonetic similarity, which can be characterized based on the number of distinctive features shared by two phones. PWLD  captures, for example, that the German word \textit{sicher} \mbox{\textipa{/z \textsc{I} \c{c} {\textturna}/}} is phonologically more similar to \textit{Becher} \textipa{/b E \c{c} {\textturna}/}  than to \textit{sitzt} \textipa{/z \textsc{I} ts t/}, even though the pairwise LD for both pairs is 2 (see Table \ref{tab:PWLD} for more examples). However, we make three adaptations to the original PWDL to make it suitable for our study: (1) we represent every phone in our inventory as a discrete, multi-valued feature vector based on the PHOIBLE \cite{moran2014phoible} feature set, (2) we compute the substitution cost between phones as the Hamming distance between their feature vector representations, and (3) we set the deletion and insertion cost to 0.5 which is equivalent to the maximum possible substitution cost. 




\section{Evaluation Tasks}

\subsection{Acoustic Word Discrimination}
The word discrimination task mainly evaluates the ability of a model to determine whether or not two given speech segments correspond to the same word type. We define this task as a segment-level retrieval task \cite{muller2015fundamentals}: given a query segment $\overline{\boldsymbol{q}}$ and a candidate set of $k$ word segments $\mathcal{S}= {\{\overline{\boldsymbol{s}}_1, ..., \overline{\boldsymbol{s}}_k\}}$, the goal is to rank segments in $\mathcal{S}$ in such a way that those segments corresponding to the same word type as the query $\overline{\boldsymbol{q}}$ are highly ranked. To this end, a vector-based search index is built by mapping each word segment in $\mathcal{S}$ into an embedding. Then, the cosine similarity between the embedding of the query $\overline{\boldsymbol{q}}$ and each embedding in the search index is computed which yields a ranked list, or an ordering $\mathcal{R}_{\boldsymbol{\theta}}$, of segments based on the cosine similarity score. The average precision metric is used to evaluate the quality of the ordering for a single query as 
\begin{equation}
    \text{AP} = \frac{1}{|\mathcal{S}_{\boldsymbol{q}}|}\sum_{r = 1}^{k}{P_{\boldsymbol{q}}(r) \times \mathcal{I}_{\boldsymbol{q}}(r)}
\end{equation}
where $\mathcal{S}_{\boldsymbol{q}}$ are the segments in $\mathcal{S}$ that correspond to the query $\overline{\boldsymbol{q}}$,  $P_{{\boldsymbol{q}}}(r)$ is the precision at rank $r$, and $\mathcal{I}_{\boldsymbol{q}}(r)$ is a relevance function such that $\mathcal{I}_{\boldsymbol{q}}(r) = 1$ if the segment at rank $r$ corresponds to the same word type as the query, or   $\mathcal{I}_{{\boldsymbol{q}}}(r) = 0$ otherwise. The arithmetic average over all AP values in the test set yields the mean average precision (mAP) metric.


\subsection{Word Phonological Similarity}
To assess whether the emerging AWE space captures word-form similarity, we propose the word phonological similarity task. We argue that an evaluation based on phonological similarity will be more insightful to understand the impact of the model architecture and learning objective on the emergent embedding space. This evaluation task works as follows: given a  query segment $\overline{\boldsymbol{q}}$ and a search index over the candidate set $\mathcal{S}$, two ranked lists, or orderings,  are produced: (1)  an ordering $\mathcal{R}_{\boldsymbol{\theta}}$ based on the cosine similarity between the AWEs, and (2) an ordering $\mathcal{R}_{\boldsymbol{\phi}}$ based on phonological similarity with the PWLD introduced in \S3. To measure the degree of agreement between the two orderings $\mathcal{R}_{\boldsymbol{\theta}}$ and $\mathcal{R}_{\boldsymbol{\phi}}$, we use Kendall's $\tau$, which is a measure of rank correlation between two ordinal variables \cite{lapata2006automatic}. For each query segment $\overline{\boldsymbol{q}}$ in the test set, Kendall's $\tau$ is computed as 

\begin{equation}
    \tau = 1 - \frac{2 \times \delta(\mathcal{R}_{\boldsymbol{\theta}}, \mathcal{R}_{\boldsymbol{\phi}})}{0.5 \times k(k-1)}
\end{equation}
where $\delta(\mathcal{R}_{\boldsymbol{\theta}}, \mathcal{R}_{\boldsymbol{\phi}})$ is the minimum number of adjacent transpositions needed to bring $\mathcal{R}_{\boldsymbol{\theta}}$ to $\mathcal{R}_{\boldsymbol{\phi}}$. Kendall's $\tau$ coefficient takes values between $1.0$ (identical ranks) and $-1.0$ (reverse ranks), while $0$ indicates no association between the two orderings. Note that Spearman's correlation is not an appropriate metric for this task, as opposed to  Kendall's $\tau$ which is designed to handle tied rankings that occur when the PWLD gives the same phonological distance for acoustic segments corresponding to the same word type.

\section{Experiments}

\subsection{Experimental Data}

\begin{table}[]
\centering
\caption{Word-level statistics of our experimental data.}
\label{tab:data-stats}
\begin{tabular}{@{}cccccc@{}}
\toprule
\multirow{2}{*}{} & \multicolumn{3}{c}{\#segments per split} & \multirow{2}{*}{\begin{tabular}[c]{@{}c@{}}\#phones\\ ($\mu \pm \text{std}$)\end{tabular}} & \multirow{2}{*}{\begin{tabular}[c]{@{}c@{}}duration\\ ($\mu \pm \text{std}$)\end{tabular}} \\ \cmidrule(lr){2-4}
 & train & valid & test &  &  \\ \midrule
German & 45886 & 7452 & 9964 & 6.8 $\pm$ 2.2 & .46 $\pm$ 0.2 \\
Czech & 68596 & 9244 & 11626 & 6.7 $\pm$ 2.5 & .50 $\pm$ 0.2 \\ \bottomrule
\end{tabular}
\end{table}

The data in our study is drawn from the GlobalPhone speech database for German and Czech \cite{schultz2013globalphone}. We choose these two languages due to their predictable grapheme-to-phoneme (G2P) mapping and the availability of high quality G2P tools. We use the Montreal Force Aligner \cite{mcauliffe2017montreal} to obtain time-aligned spoken word segments. Each acoustic word segment is parametrized  as a sequence of 39-dimensional Mel-Frequency Spectral Coefficients (MFSCs) where frames are extracted over intervals of 25ms with 10ms overlap. Table \ref{tab:data-stats} shows summary statistics of our experimental data.


\subsection{Architectures and Hyperparameters}

\textbf{CNN Acoustic Encoder.} \hspace{0.15cm} We employ a 3-layer temporal convolutional network (1D-CNN) with 256, 512, and 1024 filters and widths of 16, 32, and 48 for each layer and keep stride step at 1. Following each convolutional operation, we apply batch normalization, ReLU non-linearity, and dropout. We apply average pooling to downsample the representation at the end of the convolution block, which yields a 1024-dimensional AWE.\vspace{0.1cm} %
\noindent
\textbf{RNN Acoustic Encoder.} \hspace{0.15cm} We employ a 2-layer bidirectional Gated Recurrent Unit (BGRU) with a hidden state dimension of 512, which yields  a 1024-dimensional AWE. We apply layer-wise dropout with a probability tuned over $\{0.0, 0.2, 0.4\}$. \vspace{0.1cm}

\noindent
\textbf{Training Details.} \hspace{0.15cm} All models in this study are trained for 100 epochs with a batch size of 256  using the ADAM optimizer \cite{DBLP:journals/corr/KingmaB14} and an initial learning rate (LR) of 0.001. The LR is reduced by a factor of 0.5 if the mAP on the validation set does not improve for 10 epochs. The epoch with the best validation performance during training is used for evaluation on the test set. \vspace{0.1cm}

\noindent
\textbf{Implementation.} \hspace{0.15cm}  We build our models using PyTorch \cite{paszke2019pytorch}  and use FAISS \cite{JDH17} for efficient similarity search. Our code is publicly available on GitHub\footnote{\url{https://github.com/uds-lsv/AWEs_phon_sim}}. 

\subsection{Experimental Results}

\begin{table*}[ht]
\centering
\caption{The results of our experiments for both tasks: word discrimination (mAP) and phonological similarity ($\overline{\tau}$).   } 
\label{tab:results}
\begin{tabular}{@{}cclcclcc@{}}
\toprule
\multirow{2}{*}{\begin{tabular}[c]{@{}c@{}}Encoder\\ Architecture\end{tabular}} & \multirow{2}{*}{\begin{tabular}[c]{@{}c@{}}Learning \\ Objective\end{tabular}} &  & \multicolumn{2}{c}{\textsc{German}}      
&  & \multicolumn{2}{c}{\textsc{Czech}}                     \\ \cmidrule(l){4-8} 
&                                                                                &  & mAP $\pm$ std              & $\overline{\tau}$ $\pm$ std             &  & mAP $\pm$ std               & $\overline{\tau}$ $\pm$ std              \\ \midrule
\multirow{4}{*}{ \begin{tabular}[c]{@{}c@{}}Convolutional\\ (1D-CNN)\end{tabular} } & \textsc{\textsc{PhoneDetect}} &  & 0.552 $\pm$ 0.33 & 0.043 $\pm$ 0.17 &  & 0.669 $\pm$ 0.31 & 0.080 $\pm$ 0.14 \\
 & \textsc{Word2Phones } &  & 0.602 $\pm$ 0.33 & \text{0.127 $\pm$ 0.19} &  & 0.692 $\pm$ 0.31 & 0.150 $\pm$ 0.16 \\
 & \textsc{\textsc{\textsc{Siamese}} (w/ Rand Neg)} &  & 0.621 $\pm$ 0.33 & 0.126 $\pm$ 0.12 &  & 0.731 $\pm$ 0.30 & \text{0.176 $\pm$ 0.10} \\
 & \textsc{\textsc{\textsc{Siamese}} (w/ Hard Neg)} &  & \text{0.719 $\pm$ 0.32} & 0.074 $\pm$ 0.08 &  & \text{0.823 $\pm$ 0.27} & 0.093 $\pm$ 0.06 \\ \midrule
\multirow{4}{*}{\begin{tabular}[c]{@{}c@{}}Recurrent\\ (BGRU)\end{tabular}} & \textsc{\textsc{PhoneDetect}} &  & 0.652 $\pm$ 0.31 & \textbf{0.237 $\pm$ 0.14} &  & 0.730 $\pm$ 0.31 & 0.203 $\pm$ 0.11 \\
 & \textsc{Word2Phones} &  & 0.692 $\pm$ 0.32 & 0.181 $\pm$ 0.15 &  & 0.796 $\pm$ 0.28 & \textbf{0.226 $\pm$ 0.13} \\
 & \textsc{\textsc{\textsc{Siamese}} (w/ Rand Neg)} &  & 0.668 $\pm$ 0.34 & 0.148 $\pm$ 0.08 &  & 0.748 $\pm$ 0.31 & 0.153 $\pm$ 0.06 \\
 & \textsc{\textsc{\textsc{Siamese}} (w/ Hard Neg)} &  & \textbf{0.757 $\pm$ 0.32} & 0.044 $\pm$ 0.05 &  & \textbf{0.842 $\pm$ 0.27} & 0.077 $\pm$ 0.04 \\ \bottomrule
\end{tabular}

\end{table*}

Our results are summarized in Table \ref{tab:results} for both evaluation tasks. The word discrimination task is measured by the mAP metric while the word phonological similarity task is measured by the mean of Kendall's $\tau$ rank correlation coefficients ($\overline{\tau}$).

\noindent
\textbf{Acoustic Word Discrimination.} \hspace{0.15cm} From the mAP values reported in Table \ref{tab:results}, one can make two high-level observations: (1) all models outperform our classifier-based \textsc{PhoneDetect} baseline, which is trained to detect phone $n$-grams in the acoustic segment, for both languages and architectures, and (2) recurrent models outperform their convolutional counterparts, which is consistent with the findings reported in the literature \cite{settle2016discriminative}. However, we also observe that the performance of the \textsc{Siamese} models, which are explicitly trained to minimize the cosine distance between segments of the same word type, largely depends on the negative sampling strategy. For example, in the case of recurrent \textsc{\textsc{Siamese}} models, we observe a relative improvement in the mAP score up to 13.32\% for German and 12.57\% for Czech when applying semi-hard negative sampling. Note that the \textsc{Siamese} recurrent models did not outperform the \textsc{Word2Phones} recurrent models when the contrastive negative samples were chosen randomly from the mini-batch. These findings highlight the importance of negative sampling in training AWEs with contrastive objectives, which is a matter that has not been previously investigated to the best of our knowledge. We conclude that models trained with objectives that explicitly optimize the distance in the AWE space outperform other models that lack this objective on word discrimination, especially if the negative samples are chosen with a challenging criterion.

\noindent
\textbf{Word Phonological Similarity.} \hspace{0.15cm} In this evaluation, we observe a positive correlation between the distance in the embedding space and phonological distance for all models as indicated by the positive values of the $\overline{\tau}$ metric. Nevertheless, the correlation seems to be either weak or moderate in the best cases. Moreover, we observe that both the learning objective and encoder architecture have a considerable impact on the extent to which the embedding space captures phonological similarity. For example, the convolutional \textsc{PhoneDetect} models show some of the lowest correlation scores (German $\overline{\tau}$ = 0.043 and Czech $\overline{\tau}$ = 0.080), while their recurrent counterparts show some of the highest correlation scores (German $\overline{\tau}$ = 0.237 and Czech $\overline{\tau}$ = 0.203), even their training objective remains unchanged. These findings indicate that convolutional encoders may behave like shallow pattern detectors when trained as a classifier while recurrent encoders tend to preserve the temporal structure of the acoustic input in their representation. Another observation that we find surprising is the poor performance of the \textsc{\textsc{Siamese}} models on this task given that they outperform the other models on word discrimination. Overall, recurrent models which are trained with symbolic grounding (namely \textsc{PhoneDetect} and \textsc{Word2Phones}) are better at capturing word-form similarity compared to their convolutional counterparts on the one hand, and the recurrent \textsc{Siamese} models on the other.

\section{Discussion and Conclusion}
Although the vast majority of previous work has been driven by the engineering applications of AWEs, there is a growing scientific interest in using deep neural networks as cognitive models of (human) speech processing \cite{MatusevychSKFG20, matusevych2021phonetic, magnuson2020earshot, mayn-etal-2021-familiar}. Therefore, we argue that this cognitively motivated direction requires us to take a closer look at the embedding space and examine the degree to which we can rely on the emergent distance as an estimate of (perceptual) dissimilarity between linguistic units. In this paper, we take a step in this direction and conduct a set of experiments where we keep the training conditions for each model fixed and systematically study the impact of the architecture (convolutional and recurrent) and learning objective (classification, phonological decoding, and contrastive objectives) on the AWEs' performance using two evaluation tasks: acoustic word discrimination and word phonological similarity.


Our experiments demonstrate that while contrastive objectives yield AWEs with strong discriminative performance, they fail to reflect the phonological distance between word-forms, especially compared to AWEs which are trained with symbolic grounding (i.e., phone sequences corresponding to words). We hypothesize that the contrastive objective emphasizes word separability in the embedding space which hinders the ability of the emerging distance to reflect word similarity. Moreover, our experiments show a consistent trend with recurrent models outperforming their convolutional counterparts in both evaluation tasks, which we attribute to the ability of recurrent DNNs to model the temporal nature of speech. In conclusion, our experimental findings highlight the necessity for more diverse evaluation schemes when working with AWEs to investigate the degree to which they produce human-like errors. Furthermore, our work can be extended by analyzing the correlation between embedding distances and human perceptual similarity judgments.



\section{Acknowledgements}
We thank the anonymous reviewers for their constructive comments. This research is funded by the Deutsche Forschungsgemeinschaft (DFG, German Research Foundation), Project ID 232722074 -- SFB 1102.

\newpage 

\bibliographystyle{IEEEtran}

\bibliography{paper}


\end{document}